\documentclass[10pt,twocolumn,letterpaper]{article}

\usepackage{cvpr}              
\usepackage{comment}

\usepackage{kotex}
\usepackage{multirow}
\usepackage{tikz}
\usepackage{amssymb}

\newcommand{\redtinyb}[1]{\textbf{\scriptsize \color{red} #1}}
\newcommand{\redtiny}[1]{{\scriptsize \color{red} #1}}

\newcommand{\grayc}{\cellcolor[HTML]{C0C0C0}}
\newcommand{\graycb}[1]{\cellcolor[HTML]{C0C0C0}{\textbf{#1}}}

\definecolor{cvprblue}{rgb}{0.21,0.49,0.74}
\usepackage[pagebackref,breaklinks,colorlinks,citecolor=cvprblue]{hyperref}

\def\paperID{4780} 
\def\confName{CVPR}
\def\confYear{2025}
\usepackage[accsupp]{axessibility}
\usepackage{kotex}
\usepackage[subtle,mathdisplays=normal,wordspacing=normal]{savetrees}

\title{BIGS: Bimanual Category-agnostic Interaction Reconstruction \\from Monocular Videos via 3D Gaussian Splatting}

\author{Jeongwan On \quad Kyeonghwan Gwak \quad Gunyoung Kang \\\quad Junuk Cha \quad Soohyun Hwang \quad Hyein Hwang \quad Seungryul Baek \vspace{2mm} \\ UNIST, South Korea
}

\usepackage{float}

\def\ie{\emph{i.e}\onedot,\xspace} 

\begin{document}

\twocolumn[{
\maketitle

%
\begin{center}
    \captionsetup{type=figure}
    \includegraphics[width=\textwidth]{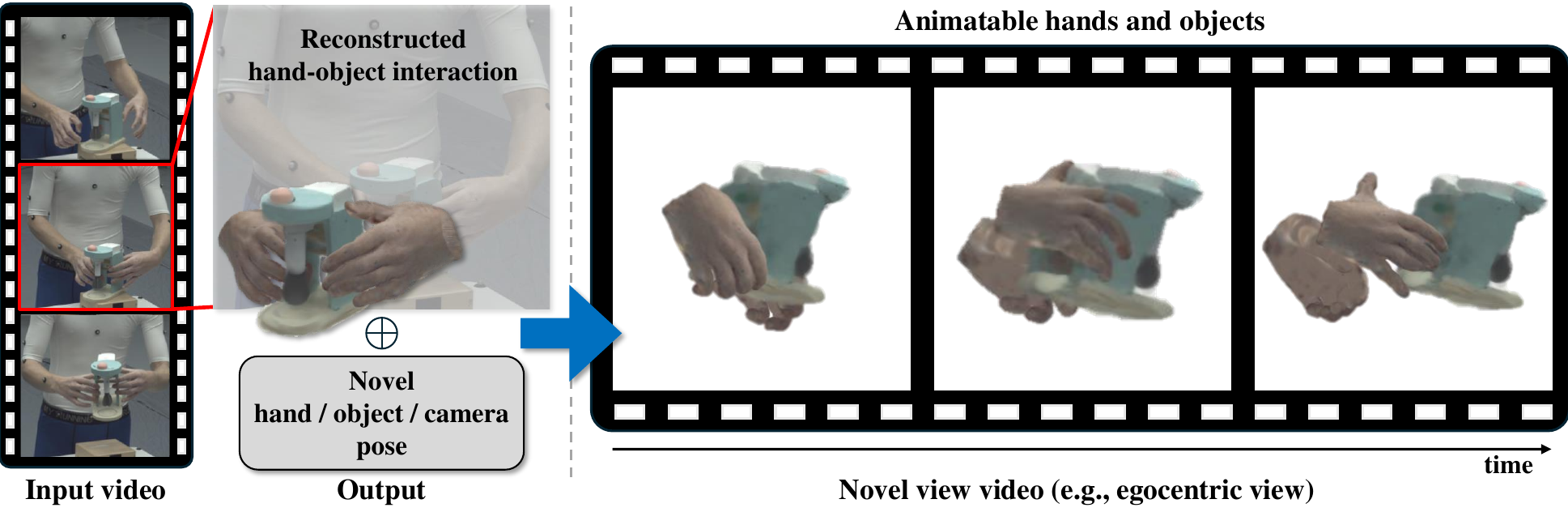}
    \captionof{figure}{Our approach reconstructs 3D Gaussians of bimanual category-agnostic interactions from a monocular video, where the two hands interact with an unknown object. Even with limited observations, our method reliably builds the 3D Gaussians in this scenario and  
    once 3D Gaussians are built, our method can be used to render new videos with novel poses of hand, object and camera (\ie view).
    }
    \label{fig:fig1}
\end{center}
}]

\begin{abstract}

Reconstructing 3Ds of hand-object interaction (HOI) is a fundamental problem that can find numerous applications. Despite recent advances, there is no comprehensive pipeline yet for bimanual class-agnostic interaction reconstruction from a monocular RGB video, where two hands and an unknown object are interacting with each other. Previous works tackled the limited hand-object interaction case, where object templates are pre-known or only one hand is involved in the interaction. The bimanual interaction reconstruction exhibits severe occlusions introduced by complex interactions between two hands and an object. To solve this, we  first introduce BIGS (Bimanual Interaction 3D Gaussian Splatting), a method that reconstructs 3D Gaussians of hands and an unknown object from a monocular video. To robustly obtain object Gaussians avoiding severe occlusions, we leverage prior knowledge of pre-trained diffusion model with score distillation sampling (SDS) loss, to reconstruct unseen object parts. For hand Gaussians, we exploit the 3D priors of hand model (i.e., MANO) and share a single Gaussian for two hands to effectively accumulate hand 3D information, given limited views. To further consider the 3D alignment between hands and objects, we include the interacting-subjects optimization step during Gaussian optimization. Our method achieves the state-of-the-art accuracy on two challenging datasets, in terms of 3D hand pose estimation (MPJPE), 3D object reconstruction (CDh, CDo, F10), and rendering quality (PSNR, SSIM, LPIPS), respectively.

\end{abstract}
\section{Introduction}
\label{sec:intro}

In our daily life, we constantly interact with objects using our hands. Accurately modeling hand-object interaction is essential for numerous applications such as virtual reality (VR), augmented reality (AR), remote robot control, and enabling more natural and intuitive human-computer interaction (HCI). Consequently, there have been many works that deal with the hand-object interaction (HOI) reconstruction problem~\cite{juchacvpr2024,elkhan2025qort,fan2024hold, pokhariya2024manus,chen2023hand,hasson2021towards,karunratanakul2020grasping,ye2022hand,ye2023diffusion,chen2023gsdf,qi2020unsupervised,hampali2022keypoint,hasson2019learning,jian2023affordpose,hampali2020honnotate,fan2023arctic,Faneccv2024,cho2023transformer,armagan2020measuring,baek2020weakly,garcia2018first,tekin2019h,kwon2021h2o,chaocvpr2021}. However, each method exhibits its own limitations: many existing HOI reconstruction methods rely on pre-scanned templates for objects~\cite{juchacvpr2024,elkhan2025qort,fan2023arctic,Faneccv2024,cho2023transformer,armagan2020measuring,garcia2018first,tekin2019h,kwon2021h2o,hampali2020honnotate,chaocvpr2021}. Although these approaches effectively use the prior information, the number of objects is typically restricted to 10 to 20 specific object shapes, which makes it challenging to extend them in the real-world scenarios having a wide variety of objects. Recently, to address the issue, Fan~\etal~\cite{fan2024hold} proposed HOLD, which successfully achieved the category-agnostic 3D reconstruction using signed distance function (SDF). However, as a neural radiance field (NeRF)~\cite{mildenhall2020nerf}-based method, HOLD~\cite{fan2024hold} requires a considerable amount of time for fitting and rendering. In another approach, Pokhariya~\etal~\cite{pokhariya2024manus} introduced a method named MANUS, which employs the 3D Gaussian Splatting (3DGS)~\cite{kerbl3Dgaussians} for hand-object interaction reconstruction. As the 3DGS-based method, MANUS achieves much faster fitting and rendering speed compared to  NeRF-based methods~\cite{fan2024hold}. However, it cannot be applied in situations where only a monocular video is provided, as it requires multi-viewed images for a particular scene.

Another important aspect is that most category-agnostic methods~\cite{fan2024hold,pokhariya2024manus} limit their scenarios where only one hand interacts with the object~\cite{fan2024hold,pokhariya2024manus,yuan2023rlipv2,chen2021joint,zhang2021single,tzionas20153d,karunratanakul2020grasping,cao2021reconstructing}. When two hands are involved, occlusion patterns of two hands and an object become dramatic and hand-object pixels cannot be properly reconstructed due to the severe occlusions. Therefore, we need to develop a dedicated scheme to explicitly reconstruct such missing pixels. Also, the recent whole body reconstruction work~\cite{moon2024exavatar}, even equipped with hand models~\cite{MANO:SIGGRAPHASIA:2017}, cannot reliably operate on 3D reconstruction of complex hand-object interactions, since images in this domain frequently miss the whole bodies; while involving frequent contacts between two hands and an object. This leads to a significant accuracy drop. These challenges indicate the necessity of developing the dedicated pipeline that effectively addresses the unique characteristic of bimanual category-agnostic interaction reconstruction.

In this paper, to overcome the limitations mentioned above, we introduce a novel method BIGS (\textbf{B}imanual \textbf{I}nteraction 3D \textbf{G}aussian \textbf{S}platting) for category-agnostic 3D reconstruction of two hands and an unknown object. Especially, BIGS builds 3D Gaussians of hands and an unknown object from monocular videos. We demonstrated that the resultant 3D reconstruction becomes far more accurate than 
hand and object Gaussians which are initialized from Transformer-based hand pose estimator~\cite{pavlakos2024reconstructing} and structure-from-motion method~\cite{schoenberger2016mvs}, respectively. Furthermore, once 3D Gaussians are built, they are used to render new scenes with novel poses of hand, object and camera, as illustrated in Fig.~\ref{fig:fig1}.

We propose the 3D Gaussian splatting-based 3D reconstruction pipeline that first builds Gaussians for hands and objects, separately and then reflects their interactions, considering the 3D contacts. To further relieve the severe  occlusions driven by complex interactions between two hands and an object, we leverage the prior knowledge of the text-to-image (T2I) diffusion models~\cite{xu2024prompt} based on score distillation sampling (SDS) loss~\cite{poole2022dreamfusion}, when obtaining the object Gaussians. This helps us to re-generate unseen surfaces of an object with the aid of pre-trained diffusion priors~\cite{xu2024prompt}, so that it makes us to learn the 3D object Gaussians more reliably on the occluded parts. We do not involve this for hand Gaussians, since hands could be reliably reconstructed thanks to hand 3D priors available from the MANO~\cite{MANO:SIGGRAPHASIA:2017} model. By further exploiting the symmetry between the right and left hands, a single hand Gaussian is shared between two hands. This approach is efficient in reconstructing 3D hand shapes and it also helps reconstruct the occluded human hand parts, since we could reconstruct shared shapes of the hand given limited views. 

Our contribution is summarized as follows:
\begin{itemize}[label={$\bullet$}]
    \item We propose the BIGS (Bimanual Interaction 3D Gaussian Splatting) pipeline that reconstructs the 3D interactions of two hands and an unknown object based on the 3D Gaussian splatting.
    
    \item We decompose the optimization into two steps: First, we optimize hand and object Gaussians separately, to reliably reconstruct 3D shapes. Then, we optimize them jointly to further consider their interactions.
    
    \item To tackle the severe occlusions introduced by complex interactions between two hands and an object, we exploit the SDS loss with diffusion priors on object Gaussians to reconstruct unseen views of the object. For hands, we share the canonical Gaussians of the right across two hands, to accumulate hand information more effectively.
    
    \item We demonstrate the superiority of the proposed method by comparing it to existing state-of-the-arts based on ARCTIC~\cite{fan2023arctic} and HO3D~\cite{hampali2020honnotate} datasets.
\end{itemize}
\section{Related Work}
\label{sec:related}

\noindent \textbf{3D hand reconstruction.}
Traditionally, 3D hand pose estimation~\cite{vinh2015hand,xu2013efficient, baek2018augmented} has been mainly achieved based on depth images obtained from Kinect sensors~\cite{kinect}. Afterwards, RGB-based hand pose estimation~\cite{sridhar2013interactive, simon2017hand, zimmermann2017learning, panteleris2018using, iqbal2018hand,kim2021twohand, lee2023imagefree} became prevalent as RGB cameras are more convenient to use and ubiquitous compared to the depth sensors. 
After the introduction of MANO 3D hand model~\cite{MANO:SIGGRAPHASIA:2017}, many traditional studies~\cite{baek2019pushing, hasson2020leveraging, tsai20233d, zhang2021interacting, pavlakos2024reconstructing, seeber2021realistichands,spurr2021adversarial,tu2023consistent,lin2024ego2handspose,li2022nimble,zhang2020handaugment,mundra2023livehand,chen2023hand,ye2022hand,ye2023diffusion} on hand reconstruction have focused on predicting MANO's pose and shape parameters from 2D RGB images. 
Some other methods directly estimate 3D coordinates of hand mesh vertices from RGB images~\cite{Li2024HHMR,wang20243d,tu2023consistent,pemasiri2021im2mesh}, instead of exploiting MANO models. More recently, research has expanded to 3D hand reconstruction through single-hand avatars. For example, Mundra~\etal~\cite{mundra2023livehand} proposed an approach that builds the photorealistic neural implicit rendering pipeline for hands with the real-time speed. 

\noindent \textbf{3D object reconstruction.}

In early works, many studies have been leveraged voxels~\cite{maturana2015voxnet,xie2019pix2vox,wang2018pixel2mesh} and points~\cite{qi2017pointnet,spurek2021modeling,chen2023hand,karunratanakul2020grasping,fan2017point} to represent the shape of an object. However, they have limitations in expressing smooth object surfaces, as each unit moves independently. To address these issues, the mesh representation~\cite{zhu2020deep,chaocvpr2021} was proposed to express the shape of objects using 3D vertices connected by triangular faces. While this representation offers the advantage in expressing watertight object shapes, it has limitations in representing high-resolution shapes due to its finite number of vertices. Moreover, this representation is not suitable for category-agnostic 3D object reconstruction as it relies on pre-scanned templates. Recently, Fan~\etal~\cite{fan2024hold} achieved category-agnostic 3D object reconstruction without relying on pre-scanned object templates by leveraging neural radiance fields (NeRF). Unfortunately, NeRF-based methods~\cite{mundra2023livehand,chen2023hand} require a significant amount of time for both training and rendering. Taking these limitations into account, 3D Gaussian splatting (3DGS)-based approaches appeared~\cite{pokhariya2024manus,
ma2024reconstructing,zhong2024generative,yang2024gaussianobject}. Unlike NeRF-based methods, these approaches can represent an object’s shape with fewer resources; while achieving higher quality rendering results.

\noindent \textbf{3D hand-object interaction reconstruction.}
In early works, many studies on HOI reconstruction tasks primarily focused on interactions involving a single hand and an object~\cite{baek2020weakly,hasson2019learning,jian2023affordpose,hampali2020honnotate}. For example, Hasson~\etal~\cite{hasson2019learning} proposed a deep learning-based HOI reconstruction framework that jointly estimates 3D meshes of a hand and an object. Additionally, Karunratanakul~\etal~\cite{karunratanakul2020grasping} learned an implicit function representing the grasping field at the contact surface between a hand and an object to reconstruct physically plausible HOI. Recently, with the development of diffusion-based models, Ye~\etal~\cite{ye2023diffusion} enabled high-quality HOI reconstruction of novel views never seen in videos. However, these deep learning-based approaches require a significant amount of training time and large datasets to achieve high accuracy. More recently, with advancements in optimization-based rendering techniques, Pokhariya~\etal~\cite{pokhariya2024manus} achieved high-quality HOI reconstruction in a very short time using 3D Gaussian splatting; however, it requires multi-viewed images for a particular scene. To reconstruct hand-object interactions from a monocular video, Fan~\etal~\cite{fan2024hold} introduced a NeRF-based pipeline and obtained the state-of-the-art 3D reconstruction result for hand-object interactions. Our framework further achieved the task of reconstructing hand object interactions involving two hands and an unknown object from a monocular video, which is more challenging.
\section{Method}
\label{sec:method}


\begin{figure*}[t]
  \centering
\includegraphics[width=1.0\linewidth]{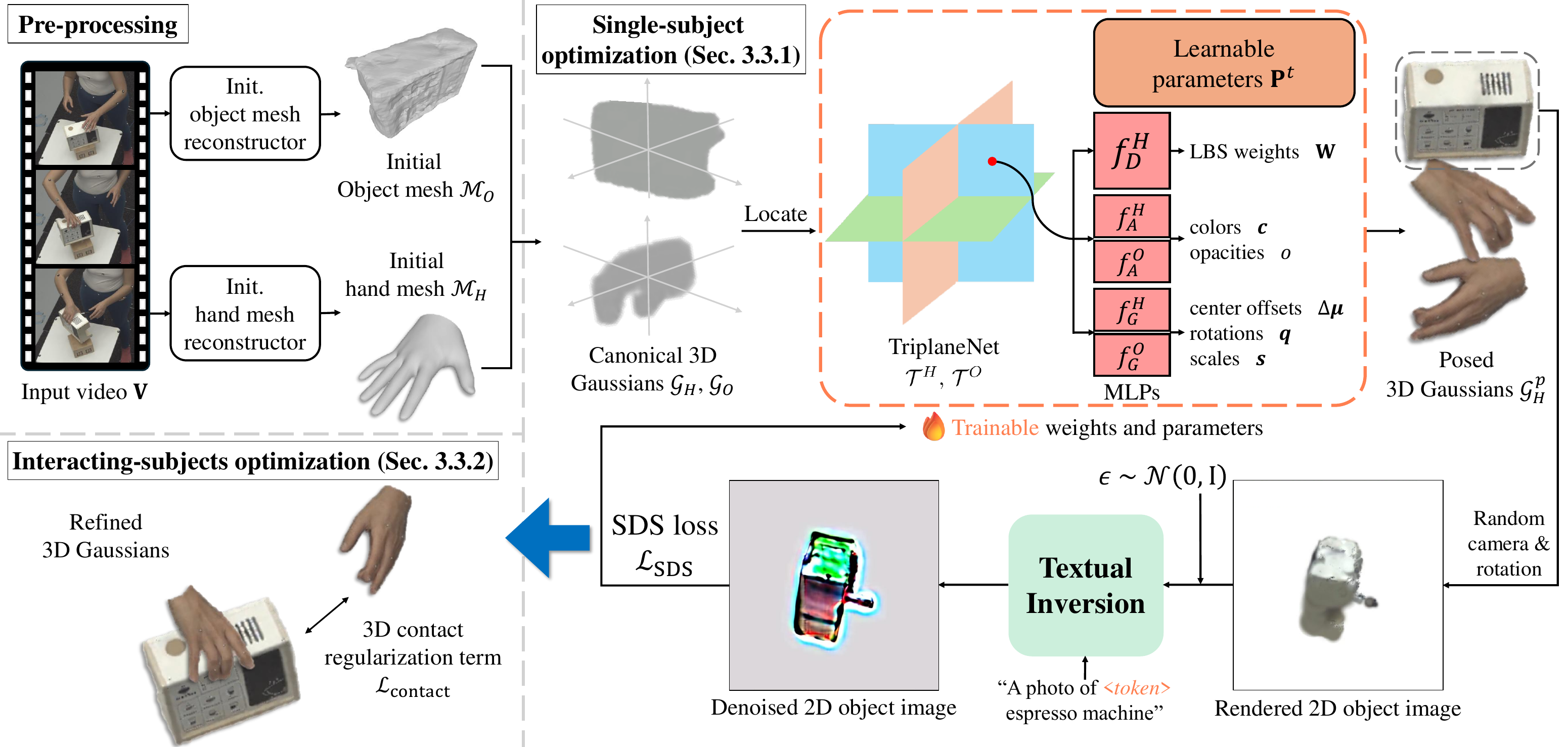}

   \caption{\textbf{Overview of the BIGS pipeline}. Initial hand meshes $\mathcal{M}_H$ and initial object meshes $\mathcal{M}_O$ are reconstructed in pre-processing step (See supplemental for details). Afterwards, the `single-subject optimization' step optimizes 3D Gaussians for hands $\mathcal{G}_H$ and objects $\mathcal{G}_O$ in the canonical space: TriplaneNet~$\{\mathcal{T}^H,\mathcal{T}^O\}$, MLPs~$\{f^H_D, f^H_A, f^O_A, f^H_G, f^O_G\}$ and learnable parameters $\mathbf{P}^t$ are updated using the Eq.~\ref{eq:single_optim}. Subsequently, the `interacting-subjects optimization' step further reflects contacts between hands and objects, and refines initial hand Gaussians $\mathcal{G}_H$ using the Eq.~\ref{eq:joint_optim}.}
   \label{fig:fig3}
\end{figure*}

This section introduces our pipeline to robustly achieve the 3D reconstruction of two hands and an unknown object. From a given input video $\mathbf{V}=\{I_t\}_{t=1}^T$ and the corresponding camera parameters $\{\mathbf{C}_t\}_{t=1}^T$ obtained from~\cite{sarlin2019coarse}, we output the 3D Gaussian splats for hands $\mathcal{G}^H$ and objects $\mathcal{G}^O$. Hand Gaussians are optimized only for the right hand and shared for two hands. Via the scheme, two-hands information could be effectively accumulated in the same canonical space by flipping the single hand Gaussians. We observed that this is effective for relieving the severe occlusions driven by bimanual class-agnostic interactions. To improve 3D Gaussians of the unknown objects, we employ the text-to-image (T2I) diffusion model~\cite{xu2024prompt} with the score distillation sampling (SDS) loss~\cite{poole2022dreamfusion}. This helps us to reliably reconstruct complete 3D object shapes even with limited viewpoints and severe occlusions. We further propose several loss functions to obtain good 3D reconstruction accuracy for the hand-object interaction, and high-fidelity rendered images. Our entire pipeline is illustrated in Fig.~\ref{fig:fig3} and we will detail each component in the remainder of this section.

\subsection{3D Gaussian splatting}

3D Gaussian splatting~\cite{kerbl3Dgaussians} is an explicit 3D representation composed of a set of 3D Gaussians, where the $i$-th Gaussian is defined by
parameters $\{ \boldsymbol{\mu}_i, \mathbf{s}_i, \mathbf{q}_i, o_i, \mathbf{c}_i \}$ where $\boldsymbol{\mu}_i\in\mathbb{R}^3$, $\mathbf{s}_i\in\mathbb{R}^3$, $\mathbf{q}_i\in SO(3)$, $o_i$ and $\mathbf{c}_i$ denote a Gaussian center, a scaling factor, a rotation quaternion, an opacity value and a color, respectively. For a 3D point cloud $\mathbf{x}$, its weight is obtained as follows:
\begin{eqnarray}    \mathcal{G}_i(\mathbf{x}) = e^{-\frac{1}{2}(\mathbf{x}-\boldsymbol{\mu}_i)^T\boldsymbol{\Sigma}^{-1}_i(\mathbf{x}-\boldsymbol{\mu}_i)}
\end{eqnarray}
where a covariance matrix $\boldsymbol{\Sigma}_i$ is calculated based on $\mathbf{s}_i$ and $\mathbf{q}_i$. The color of a pixel is calculated as:
\begin{eqnarray}
     C(\mathbf{x}_p) &=& \sum_{i \in N} c_i \Tilde{o}_i \prod^{i-1}_j (1-\Tilde{o}_j),\label{eq:gaussian rendering1}
\end{eqnarray}
where $\Tilde{o}_i = o_i \mathcal{G}^{proj}_i(\mathbf{x}_p)$ 
and 2D Gaussian $\mathcal{G}^{proj}_i$ is projected from the 3D Gaussian $\mathcal{G}_i$ onto the image space.

\subsection{Bimanual interaction 3D Gaussian splatting}
\label{sec:3_2}

We propose the bimanual interaction 3D Gaussian splatting (BIGS) that inherits the original 3D Gaussian splatting~\cite{kerbl3Dgaussians}; while having a few novel schemes to properly tackle the bimanual category-agnostic interaction reconstruction task.

\noindent \textbf{Pre-processing.} Given an input video $\mathbf{V}$ having $T$ frames, we first extract a hand mesh $\mathcal{M}_H^t=\{\mathcal{M}_L^t, \mathcal{M}_R^t\}$ and an object mesh $\mathcal{M}_O^t$ for the $t$-th frame, where $t\in[1,T]$. The vertices of the obtained hand mesh $\mathcal{M}_H^t$ and the vertices of the obtained object mesh $\mathcal{M}_O^t$ are further used as the initial Gaussian locations in our BIGS.
To obtain hand meshes, we first use the hand mesh regressor~\cite{pavlakos2024reconstructing} that estimates MANO pose parameters $\theta_L^t, \theta_R^t\in\mathbb{R}^{45}$ of left and right hands, and the shape parameters $\beta\in\mathbb{R}^{10}$, where the shape parameters $\beta$ are shared across all frames, by fixing it as the first frame's right shape parameter, as they are same person's hands. Global orientation $\{\Phi_L^t, \Phi_R^t\}\subset SO(3)$ and translation $\{\Gamma_{L}^t, \Gamma_{R}^t\}\subset \mathbb{R}^{3}$ parameters
are also obtained from~\cite{pavlakos2024reconstructing}. For the object meshes, we involve the structure-from-motion (SfM) approach~\cite{sarlin2019coarse} to obtain the point cloud of the object and its rotation $\Phi^{t}_{O}\in SO(3)$ and translation $\Gamma^t_O\in\mathbb{R}^3$ parameters from the $t$-th frame. Then, we obtain the aligned hand-object meshes following~\cite{fan2024hold}. See supplemental for more details.

\noindent \textbf{Gaussians for two-hands.} The hand Gaussians are constructed in the canonical space. The feature $\mathbf{t}\in\mathbb{R}^d$ is extracted from the hand TriplaneNet $\mathcal{T}^H$~\cite{triplane_iccv23} by interpolating the center of the $i$-th Gaussian $\boldsymbol{\mu}_i$. Then, taking the feature $\mathbf{t}$, three multi-layer perceptrons (MLPs) estimate the parameters of the $i$-th Gaussian: the hand appearance MLP $f^H_A$ predicts a color $\mathbf{c}_i$ and an opacity $o_i$; the hand geometry MLP $f^H_G$ predicts an offset to the center location $\Delta \boldsymbol{\mu}_i\in\mathbb{R}^3$, rotation $\mathbf{q}_i\in SO(3)$ and scale $\mathbf{s}_i\in\mathbb{R}^3$; the hand deformation MLP $f^H_D$ predicts the LBS weight $\mathbf{W}_i\in\mathbb{R}^{K}$, where $K$ denotes the number of hand joints.

The center of the $i$-th hand Gaussian in the canonical space $\boldsymbol{\mu}_i^c$ is transformed to the posed location $\boldsymbol{\mu}_i^p$, by applying the linear blend skinning (LBS) weights $\mathbf{W}\in\mathbb{R}^{K}$ of the MANO model~\cite{MANO:SIGGRAPHASIA:2017} whose element $\mathbf{W}^{k}(\mathbf{v})$ encodes the weight between the $k$-th joint to the vertex $\mathbf{v}$ on the hand mesh, as follows:
\begin{eqnarray}
\boldsymbol{\mu}_i^p = \sum_{k=1}^K \mathbf{W}^k(\boldsymbol{\mu}_i^c)(\Phi_k\boldsymbol{\mu}_i^c+\Gamma_k)
\label{lbs}
\end{eqnarray}
where $\Phi_k$ and $\Gamma_k$ are rotation and translation of the $k$-th hand joint, which are computed from $\beta$ and $\theta_t$ of the MANO model, respectively. 

By flipping the center of the 3D Gaussians in the x-axis, we can map the same Gaussian to left and right hands. Therefore, we defined one single 3D Gaussian $\mathcal{G}_H$ for right hand and share it across two hands. This scheme is also useful to reason about severe occlusions driven by two hands and an unknown object interactions. Since we use the same single Guassian for two hands, we need to flip the center of Gaussian in the x-axis before calculating the Eq.~\ref{lbs} for left hands. After transforming the Gaussians in the canonical space into the posed hand Gaussians $\mathcal{G}_H^p$ for each hand, where $H\in\{L, R\}$ indicates the index for left and right handedness, we further transform the center of posed hand Gaussians to the image-coordinate hand Gaussians ${}^{c}\mathcal{G}_H$, by applying the rotation $\Phi_H^t$ and translation $\Gamma_H^t$ parameters to the posed Gaussians $\mathcal{G}_H^p$.

\noindent \textbf{Gaussians for unknown objects.} The 3D Gaussian for object $\mathcal{G}_O$ is also constructed in the canonical space. Then, the feature is extracted from the object TriplaneNet $\mathcal{T}^O$~\cite{triplane_iccv23} and then fed to two MLPs (\ie $f^O_A$, $f^O_G$) to infer parameters of the object Gaussians $\mathcal{G}_O$: The object appearnce MLP $f^O_A$ predicts color $\mathbf{c}_i$ and opacity $o_i$; the object geometry MLP $f^O_G$ predicts an offset to the center location $\Delta \boldsymbol{\mu}_i\in\mathbb{R}^3$, rotation $\mathbf{q}_i\in SO(3)$ and scale $\mathbf{s}_i\in\mathbb{R}^3$. The object Gaussians in the image-coordinate space ${}^{c}\mathcal{G}_O$ is obtained, by applying the rotation $\Phi_O^t$ and translation $\Gamma_O^t$ parameters to object Gaussians $\mathcal{G}_O$ in the canonical space.

\noindent \textbf{Rendering.} The bimanual interaction 3D Gaussians $\{\mathcal{G}_H, \mathcal{G}_O\}$ are used to render new hand-object interaction images given novel poses of hand, object and camera (\ie view). The examples are visualized in the supplemental.

\subsection{Optimization}
Our optimization step is composed of two distinct steps: (1) single-subject optimization step, (2) interacting-subjects optimization step. In the single-subject optimization step, we optimize hand Gaussians and object Gaussians separately by checking their 2D projected images consistent to the original input images. Especially, we train the hand and object Triplane-Net $\{\mathcal{T}^H, \mathcal{T}^O\}$, and MLPs $\mathbf{f}=\{f^H_A,f^H_G,f^H_D,f^O_A,f^O_G\}$. We further define the learnable parameters at the $t$-th frame as, $\mathbf{P}^t=\{\{\boldsymbol{\mu}^t_i\}_{i=1}^N, \theta_L^t, \theta_R^t, \Gamma_L^t, \Phi_L^t, \Gamma_R^t, \Phi_R^t, \Gamma_O^t, \Phi_O^t\}$ and optimize them for all frames.

To further tackle severe occlusions in bimanual class-agnostic interaction reconstruction, we involved the additional SDS loss~\cite{poole2022dreamfusion} using pre-trained stable diffusion network~\cite{xu2024prompt}. This leads us to reconstruct unseen surfaces of the 3D objects. This is only applied to the object surfaces, since we could reliably reconstruct the hands thanks to hand priors by MANO model~\cite{MANO:SIGGRAPHASIA:2017} and shared Gaussians across two hands. The mask consistency and temporal smoothness on the hand poses are further enforced.

In the interacting-subjects optimization step, we optimize the translation vector of hands $\{\Gamma^t_L, \Gamma^t_R\}_{t=1}^T$ to properly align locations of hands and objects, by considering their distance, at every frame. We will detail each step in the remainder of this subsection.

\subsubsection{Single-subject optimization step.}
\label{sec:single_step}

In this step, we optimize parameters of Triplane-Net $\mathcal{T}=\{\mathcal{T}^H, \mathcal{T}^O\}$, and parameters of MLPs $\mathbf{f}$ and learnable parameters at the $t$-th frame, $\mathbf{P}^t$, by minimizing the objective, as follows:
\begin{eqnarray}
    \underset{        \begin{smallmatrix}
            \\ \mathcal{T}, \mathbf{f}, \mathbf{P}^t
        \end{smallmatrix}
    }{\min} \mathcal{L}^t_{\text{image}} + \mathbb{I}_\text{hand}\mathcal{L}^t_{\text{hand}} + \mathbb{I}_\text{obj}\mathcal{L}^t_{\text{obj}}  \label{eq:single_optim}
\end{eqnarray}
where $\mathbb{I}_\text{hand}$ is set as $1$ for hand Gaussians and $0$ for others; while $\mathbb{I}_\text{obj}$ is set as $1$ for object Gaussians and $0$ for others. Each loss function is detailed in the following paragraphs: 

\noindent \textbf{Image consistency loss $\mathcal{L}_\text{image}$.} This loss compares the image rendered from our BIGS $\mathbf{I}_\text{pred}^t$ 
 with the original image $\mathbf{I}^t$ at the $t$-th frame. To obtain the foreground images, we obtain the foreground masks of left and right hands $\mathbf{m}_L^t$,  $\mathbf{m}_R^t$ and objects $\mathbf{m}_O^t$ via the pre-trained model~\cite{ravi2024sam2} and then multiply it with the rendered and original images as: $\mathbf{I}_\text{pred}^t\odot\mathbf{m}_H^t$ and $\mathbf{I}^t\odot\mathbf{m}_H^t$ where $H\in\{L,R,O\}$. Then, we compare two foreground regions by combining L1 loss, SSIM loss~\cite{1284395} $\mathcal{L}^t_{\text{SSIM}}$ and VGG loss~\cite{simonyan2015deepconvolutionalnetworkslargescale} $\mathcal{L}^t_{\text{VGG}}$. Rendering quality regularizers (\ie $\mathcal{L}^t_\text{color}$, $\mathcal{L}^t_\text{scale}$) are further employed from~\cite{moon2024exavatar}. Additionally, we merge foreground masks of hands and objects, and make the combined foreground mask $\bar{\mathbf{m}}^t$. Then, we propose to use a mask regularizer $\mathcal{L}^t_\text{mask}$ that prevents outlier Gaussians, by enforcing that Gaussians are generated inside the foreground mask $\bar{\mathbf{m}}^t$. The overall loss is defined as follows:
\begin{eqnarray}    \mathcal{L}^t_{\text{image}} &=& \|\mathbf{m}^t_h \odot (\mathbf{I}^t_{\text{pred}} -  \mathbf{I}^t_{\text{gt}})\|_1   + \lambda_1 \mathcal{L}^t_{\text{ssim}} + \lambda_2 \mathcal{L}^t_{\text{vgg}} \nonumber \\
&+& \lambda_3 \mathcal{L}^t_{\text{color}} + \lambda_4 \mathcal{L}^t_{\text{scale}} + \lambda_5 \mathcal{L}^t_\text{mask},  \label{eq:obj_param}
\end{eqnarray}
\noindent where $\mathcal{L}_\text{mask}=(1 - {\Bar{\mathbf{m}}}^t) \odot \mathbf{I}^t_{\text{pred}}$, $\lambda_1 = 0.2, \lambda_2 = 1.0, \lambda_3 = 0.1$ and $\lambda_4=100.0$ and $\lambda_5=10.0$. 

\noindent \textbf{Hand loss $\mathcal{L}_\text{hand}$.} This loss is proposed to train only the hand Gaussians, to enforce the smoothness of the pose parameter $\theta^t$, and smoothness of the translation parameter {$\Gamma_H^t$, where $H\in\{L, R\}$. We further regularize the LBS weight $\mathbf{W}^t$ similarly to~\cite{hugs}, by closing it to the pseudo ground-truth LBS $\mathbf{\hat{W}}^t$ that is obtained by retrieving $6$ nearest vertices on the MANO mesh and taking a distance-weighted average of LBS weights. The overall loss is defined as follows:
\begin{eqnarray}
\mathcal{L}^t_\text{hand}=\mathcal{L}^t_\text{smt} + \lambda_\text{LBS}\|\mathbf{W}^t-\mathbf{\hat{W}}^t\|_\text{F}^2
\end{eqnarray}
where $\mathcal{L}^t_\text{smt}=\| \theta^{t} - \theta^{t-1}\| ^2_2 + \sum_{H \in \{L, R \}} \| \Gamma^{t}_H - \Gamma^{t-1}_H \| ^2_2$ and $\lambda_\text{LBS}=10^3$.

\noindent{\bf Object loss $\mathcal{L}_\text{obj}$.} This loss is proposed to train only the object Gaussians, to reconstruct the occluded regions of the object. Especially, we employed the SDS loss~\cite{poole2022dreamfusion} as the object loss $\mathcal{L}^t_\text{obj}$ to provide the guidance for insufficient observations. Similar to Lee~\etal~\cite{lee2024guess}, we first involved the textual inversion~\cite{gal2023an,kumari2022customdiffusion} on the diffusion network~\cite{xu2024prompt} to get a text prompt $y$ that best describes the object. Then, we apply the diffusion-guided optimization step. 

In the textual inversion step, we learn a text prompt $y$, such as “A photo of $<\!\!token\!\!>$ $<\!\!object\!\!>$” to better represent the object that we want to reconstruct. Here, $<\!\!token\!\!>$ is a learnable embedding vector, and $<\!\!object\!\!>$ is the name of the object to be reconstructed. Additionally, to prevent the object from appearing at random locations within the generated images, we employed the PiDi boundary-conditioned ControlNet~\cite{zhang2023adding} and additionally condition the diffusion network using PiDi boundary $\mathbf{I}_\text{PiDi}$ extracted from the rendered image via~\cite{Su_2021_ICCV}. As a result, the image of the object and its PiDi boundary are fed into the T2I diffusion and ControlNet $c(\cdot)$, respectively, learning the optimal textual prompt $y$.

In the diffusion-guided optimization step, we first reconstruct the canonical object Gaussian $\mathcal{G}_O$ and create a random virtual camera $c$, which moves along a virtual sphere near $\mathcal{G}_O$. After that, we obtain the rendered foreground object image $\mathbf{I}^c$ by projecting the object Gaussians $\mathcal{G}_O$ into a random camera $c$, and use this to employ the following SDS loss:
\begin{equation}
    \nabla \mathcal{L}_{\text{SDS}} = \mathrm{E}_{t, \epsilon}[\omega(t)(\epsilon_{\phi}(z_{t}; y, c(\mathbf{I}_\text{PiDi}), t) - \epsilon)\frac{\partial z}{\partial \mathbf{I}^c}\frac{\partial \mathbf{I}^c}{\partial \mathcal{G}_O} ],
    \label{eq:sds_loss}
\end{equation}
where $t$, $\epsilon$ and $\epsilon_\phi$ denote the diffusion time step, noise and UNet in the stable diffusion~\cite{xu2024prompt}, respectively. $\omega(t)$ is the weighting function defined by the scheduler of the diffusion model.

\begin{table*}[t]
\centering
\resizebox{\textwidth}{!}{
    \begin{tabular}{c|c|cc|cccc}
    \toprule
     & & \multicolumn{2}{c|}{\textbf{Contact reconstruction}}  & \multicolumn{4}{c}{\textbf{Hand/object reconstruction}} \\ \hline
     & & CD$_l \; [cm^2] \downarrow$ & CD$_r \; [cm^2] \downarrow$ & CD$_o \; [cm^2] \downarrow$ & F10$ \; [\%] \uparrow$ & MPJPE$_l \; [mm] \downarrow$ & MPJPE$_r \; [mm] \downarrow$ \\ \hline
    \multirow{4}{*}{\begin{tabular}[c]{@{}c@{}}Two-Hand\\ Settings\end{tabular}} 
     & HLoc~\cite{sarlin2019coarse} & N/A & N/A & 4.24 & 40.57 & N/A & N/A \\
     & HaMeR~\cite{pavlakos2024reconstructing} & N/A & N/A & N/A & N/A & 27.23 & 24.37 \\
     & HOLD & 105.92 & 123.54 & 2.07 & 63.92 & 27.13 & 24.70 \\
     & \graycb{Ours*} & \graycb{46.11}~\redtinyb{-59.81} & \graycb{31.28}~\redtinyb{-92.26} & \graycb{1.28}~\redtiny{-0.79} & \graycb{83.93}~\redtinyb{+20.01} & \graycb{24.63}~\redtiny{-2.5} & \graycb{24.35}~\redtiny{-0.02} \\ 
    \bottomrule
    \end{tabular}
}
\caption{Quantitative results on ARCTIC~\cite{fan2023arctic}: Our method achieved the best accuracy on 3D reconstruction of hands ($\text{MPJPE}_l$, $\text{MPJPE}_r$), objects ($\text{CD}_o$, $\text{F}10$ ) and their contacts ($\text{CD}_l$, $\text{CD}_r$).}
\label{table:table1}
\end{table*}

\begin{table*}[t]
\centering
\resizebox{\textwidth}{!}{%
    \begin{tabular}{c|c|c|c|ccc}
    \toprule
     & & & \multicolumn{1}{c|}{\textbf{Contact reconstruction}} & \multicolumn{3}{c}{\textbf{Hand/object reconstruction}} \\ \hline
     & & & CD$_r \; [cm^2] \downarrow$ & CD$_o \; [cm^2] \downarrow$ & F10$ \; [\%] \uparrow$ & MPJPE$_r \; [mm] \downarrow$ \\ \hline
    \multirow{8}{*}{\begin{tabular}[c]{@{}c@{}}One-Hand\\ Settings\end{tabular}} 
    & \multirow{4}{*}{Full View} 
     & HOMan~\cite{hasson2021towards} & 78.2 & N/A & N/A & 32.0 \\
     & & iHOI~\cite{ye2022hand} & 75.8 & 3.8 & 41.7 & 38.4 \\
     & & DiffHOI~\cite{ye2023diffusion} & 68.8 & 4.3 & 43.8 & 32.3 \\
     & & HOLD & 11.3 & 0.4 & 96.5 & 24.2 \\
     & & \graycb{Ours*} & \graycb{11.1}~\redtiny{-0.2} & \graycb{0.3}~\redtiny{-0.1} & \graycb{96.7}~\redtiny{+0.2} & \graycb{23.9}~\redtiny{-0.3} \\ \cline{2-7}
    & \multirow{4}{*}{Limited View} 
     & & \multicolumn{1}{c|}{\textbf{Highly dynamic contacts}} & \multicolumn{3}{c}{\textbf{One side view}} \\ \cline{3-7}
     & & & CD$_r \; [cm^2] \downarrow$ & CD$_o \; [cm^2] \downarrow$ & F10$ \; [\%] \uparrow$ & MPJPE$_r \; [mm] \downarrow$ \\ \cline{3-7}
     & & HOLD & 95.63 & 3.39 & 74.90 & 24.96 \\
     & & \graycb{Ours*} & \graycb{21.70}~\redtinyb{-73.93} & \graycb{1.24}~\redtiny{-2.15} & \graycb{84.02}~\redtiny{+9.12} & \graycb{24.22}~\redtiny{-0.74} \\
    \bottomrule
    \end{tabular}
}
\caption{Quantitative results on HO3Dv3~\cite{hampali2020honnotate}: Our method achieved the best accuracy on 3D reconstruction of right hands ($\text{MPJPE}_r$), objects ($\text{CD}_o$, $\text{F}10$) and their contacts ($\text{CD}_r$). We also evaluate on the `Limited View' case, and observed that ours still reliably reconstruct 3Ds of hands, objects and their contacts; while HOLD~\cite{fan2024hold} suffers from achieving the correct 3D reconstruction, showing the significant drop compared to the `Full View' results.} 
\label{table:table2}
\end{table*}

\begin{table*}[t]
\centering
\resizebox{\textwidth}{!}{%
    \begin{tabular}{cccc|cc|cccc}
    \toprule
     & & & & \multicolumn{2}{c|}{\textbf{Contact reconstruction}} & \multicolumn{4}{c}{\textbf{Hand/object reconstruction}} \\ \hline
     w/ $\mathcal{L}_{\text{SDS}}$ & w/ $\mathcal{L}_{\text{contact}}$ & w/ $\mathcal{L}_{\text{smt}}$ & w/ $share$* & CD$_l \; [cm^2] \downarrow$ & CD$_r \; [{cm}^2] \downarrow$ & CD$_o \; [cm^2] \downarrow$ & F10$ \; [\%] \uparrow$ & MPJPE$_l \; [mm] \downarrow$ & MPJPE$_r \; [mm] \downarrow$ \\ \hline
     & & & & 96.56 & 117.72 & 1.36 & 81.78 & 27.22 & 25.08 \\
     & & & \checkmark & 96.55~\redtiny{-0.01} & 117.72~\redtiny{-0.00} & 1.36~\redtiny{-0.00} & 81.78~\redtiny{+0.00} & 27.01~\redtiny{-0.21} & 24.42~\redtiny{-0.66} \\
     & & \checkmark & \checkmark & 83.14~\redtinyb{-13.41} & 96.95~\redtinyb{-20.77} & 1.36~\redtiny{-0.00} & 81.78~\redtiny{+0.00} & 24.63~\redtinyb{-2.38} & 24.35~\redtiny{-0.07} \\
     & \checkmark & \checkmark & \checkmark & 50.78~\redtinyb{-32.36} & 35.57~\redtinyb{-61.38} & 1.36~\redtiny{-0.00} & 81.78~\redtiny{+0.00} & 24.63~\redtiny{-0.00} & 24.35~\redtiny{-0.00} \\
     \checkmark & \checkmark & \checkmark & \checkmark & \grayc{46.11}~\redtiny{-4.67} & \grayc{31.28}~\redtiny{-4.29} & \grayc{1.28}~\redtinyb{-0.08} & \grayc{83.93}~\redtinyb{+2.15} & \grayc{24.63}~\redtiny{-0.00} & \grayc{24.35}~\redtiny{-0.00} \\
    \bottomrule
    \end{tabular}
}
\caption{Ablation study on ARCTIC~\cite{fan2023arctic}: We conducted quantitative analysis on our design choices. $share$* denotes sharing the canonical 3D Gaussians of both hands.}
\label{table:table3}
\end{table*}

\subsubsection{Interacting-subjects optimization step.} 
The 3D Gaussian splatting~\cite{kerbl3Dgaussians} performs well and it generates the 3D representation well suited to the observed video $\mathbf{V}$. However, when the camera viewpoints are limited and occlusions are severe, as in our case, we observed that it suffers from reliably reconstructing 3Ds of the parts. This leads to the distortion in the 3D locations of the hands and objects, which prevents the precise contact between hands and objects. To relieve the phenomenon, we propose the additional interacting-subjects optimization step, which is able to make hand and object Gaussians more aligned each other. Especially, we propose a contact regularization term $\mathcal{L}^t_{\text{contact}}$ to encourage hand-object Gaussians to be well-contacted during the optimization:
\begin{equation}    \mathcal{L}^t_{\text{contact}} = \lambda_\text{contact}  \sum_{H \in \{L, R \}}{ || \Gamma^t_O - \Gamma^t_H ||_2 }.
  \label{eq:contact_loss}
\end{equation}
However, we tried to make the hand Gaussians not follow the object locations too closely, so we set the $\lambda_\text{contact}$ as a small value (\ie $1.0$). Finally, we find the $\Gamma_L^t$, $\Gamma_R^t$ for $t\in[1,T]$ by minimizing the objective, which is defined as follows:
\begin{equation}
    \underset{   \{\Gamma^t_L, \Gamma^t_R\} 
    }{\min} \mathcal{L}_{\text{image}} + \mathcal{L}_{\text{mask}} + \mathcal{L}_{\text{contact}}.
  \label{eq:joint_optim}
\end{equation}
\section{Experiments}
\label{sec:experiments}
In this section, we will explain our experimental setup (datasets, metrics) and analyze our results.

\begin{table}[t]
\centering
\resizebox{0.45\textwidth}{!}{%
    \begin{tabular}{c|c|ccc}
    \toprule
     &  & \multicolumn{3}{c}{\textbf{Rendering}} \\ \hline
     & & PSNR$ \; \uparrow$ & SSIM$ \; \uparrow$ & LPIPS$ \; \downarrow$ \\ \hline
    \multirow{2}{*}{\begin{tabular}[c]{@{}c@{}}Two-Hand\\ Settings\end{tabular}} & HOLD & 12.83 & 0.66 & 0.32 \\
     & \graycb{Ours*} & \graycb{24.87}~\redtinyb{+12.04} & \graycb{0.96}~\redtiny{+0.30} & \graycb{0.05}~\redtiny{-0.27} \\ \hline
\multirow{2}{*}{\begin{tabular}[c]{@{}c@{}}One-Hand\\ Settings\end{tabular}} & HOLD & 16.20 & 0.74 & 0.21 \\
     & \graycb{Ours*} & \graycb{24.51}~\redtiny{+8.31} & \graycb{0.92}~\redtiny{+0.18} & \graycb{0.07}~\redtiny{-0.14} \\
    \bottomrule
    \end{tabular}
}
\caption{\textbf{Quantitative results using rendering metrics}. Our method achieved  superior rendering quality in both two-hand and one-hand settings.}
\label{table:table4}
\end{table}

\subsection{Datasets}
\label{sec:exp_datasets}
\noindent {\bf ARCTIC Dataset.} We conducted our experiments using the ARCTIC~\cite{fan2023arctic} dataset, which provides sequences showing two hands skillfully manipulating various objects. This dataset includes video frames along with detailed 3D meshes of both hands and the objects. We followed the setting of ECCV 2024 HANDS workshop challenge~\cite{eccv2024hands}: For training set, we used data involving $9$ objects, focusing specifically on subject 3 and camera index 1. We used only sequences categorized under the `grab' action to ensure that the object remained rigid across the frames and selected $300$ frames where the hands and objects were most clearly visible.

\noindent {\bf HO3Dv3 Dataset.} We further conducted experiments using the HO3Dv3 dataset~\cite{hampali2020honnotate} that have one-hand and an object interactions. Following the setting of~\cite{fan2024hold}, we utilized $18$ available sequences for our evaluation. We call the above setting as `HO3D (Full view)'. Since videos in HO3D dataset include the frames to horizontally rotate the objects, they include the full viewpoints of the object. We on purpose sample $25\%$ of the frames in each original sequence to make the setting more challenging, by not including the full viewpoints of the object. We call this setting as `HO3D (Limited view)'. We conducted our experiments using these two datasets.

\subsection{Metric}
We used mainly three metrics for measuring hand/object reconstruction accuracy, contact reconstruction accuracy and rendering quality, in Tables~\ref{table:table1}-\ref{table:table4}. The subscripts $l$, $r$, and $o$ denote the left hand, right hand, and object, respectively.

\noindent {\bf Hand/object reconstruction.} We use the mean per joint position error (MPJPE) to measure hand pose error. For object reconstruction performance, we use Chamfer distance (CD$_o$) to measure the shape dissimilarity between the ground truth point cloud and the predicted point cloud, and the F-score, which is the harmonic mean of precision and recall, calculated with threshold of 10mm (F10), following~\cite{fan2024hold}.

\noindent {\bf Contact reconstruction.} To evaluate the 3D contact between hands and objects, we use the hand-relative Chamfer distance (CD$_l$ and CD$_r$), introduced in~\cite{fan2024hold}.

\noindent {\bf Rendering.} We use the peak signal-to-noise ratio (PSNR), structural similarity (SSIM)~\cite{1284395}, and learned perceptual image patch similarity (LPIPS)~\cite{8578166} metrics, which are most commonly used in the literature, when measuring the quality of rendered images.

\begin{figure*}[t]
  \centering
   \includegraphics[width=\linewidth]{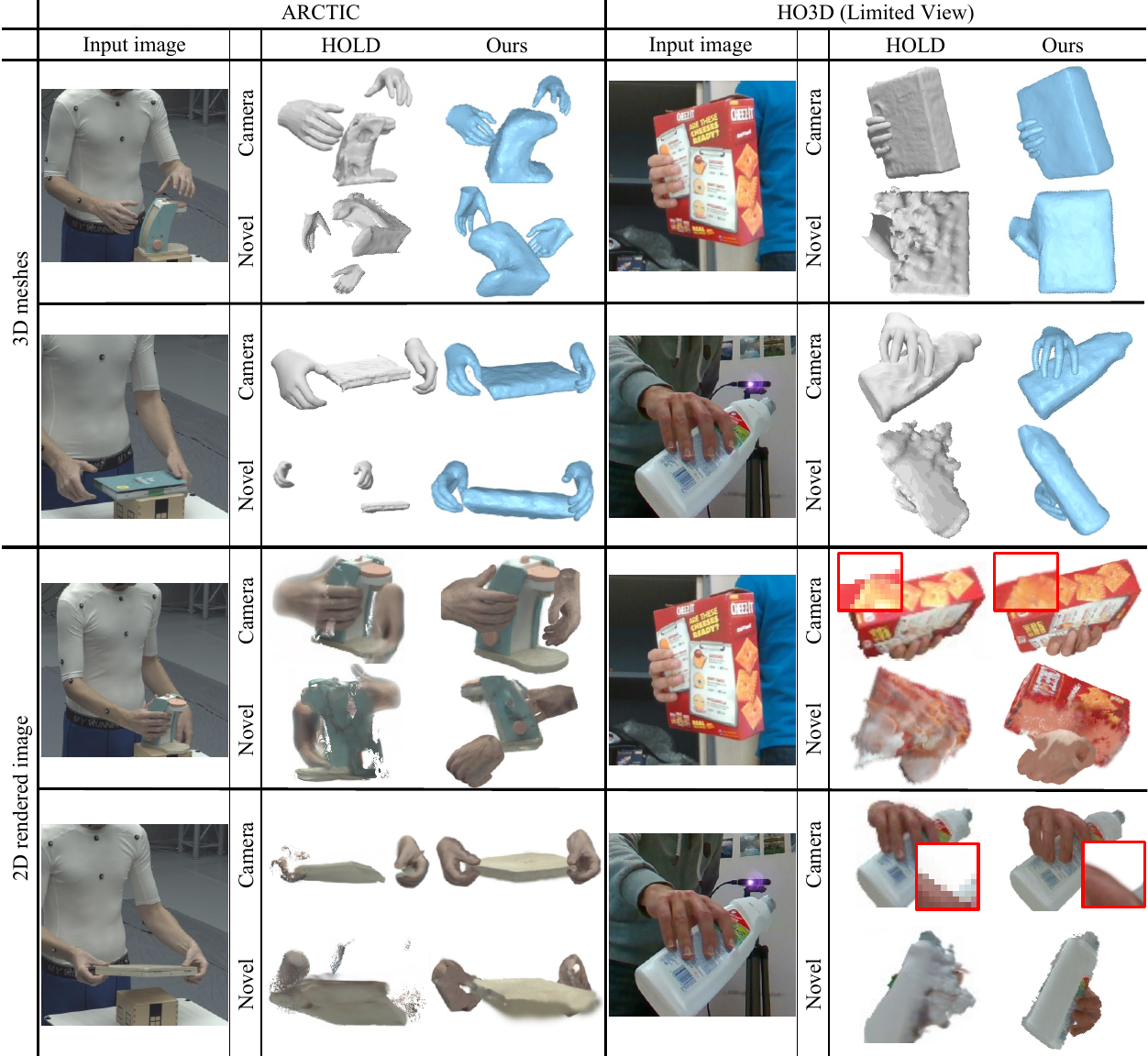}

   \caption{\textbf{Qualitative examples for 3D meshes (Rows 1, 2) and 2D rendered images (Rows 3, 4)}: Here, we exemplify 3D meshes and 2D rendered images, obtained from `HOLD'~\cite{fan2024hold} and `Ours' on ARCTIC~\cite{fan2023arctic} and HO3D~\cite{hampali2020honnotate} datasets, respectively.
   For each example, first rows visualize results in the original viewpoint and second rows visualize results in the novel viewpoint. From 3D mesh examples, we can see that `Ours' exhibits better alignment between hands and objects in the 3D space both in the camera and novel viewpoints; while the outputs of `HOLD'~\cite{fan2024hold} are not complete even in the camera viewpoint and 3D locations become completely wrong in the novel viewpoints. `Ours' exhibits cleaner rendering quality both in the camera and novel viewpoints; while the outputs of `HOLD'~\cite{fan2024hold} suffers from the noise.
   }
   \label{fig:fig3-a}
   
\end{figure*}

\subsection{Results}
\noindent \textbf{Compared algorithms.} We compared ours with several competitive baselines such as HOLD~\cite{fan2024hold}. HOLD is a NeRF-based method and utilizes a signed distance function (SDF) to reconstruct an unknown object. HaMeR~\cite{pavlakos2024reconstructing} is a Transformer-based model and utilizes Vision Transformer (ViT)~\cite{dosovitskiy2020vit} as a backbone with multiple datasets to reconstruct hand mesh. iHOI~\cite{ye2022hand} reconstructs an unknown object by utilizing hand articulation-aware coordinates and SDF. DiffHOI~\cite{ye2023diffusion} leverages a diffusion network as a prior of novel-view rendering and reconstructs 3Ds of HOI with priors and SDS loss.

\noindent \textbf{Quantitative results.} \noindent In Table~\ref{table:table1}, we evaluate our model for contact reconstruction and hand/object reconstruction quality on the ARCTIC dataset~\cite{fan2023arctic}. HOLD~\cite{fan2024hold} encourages contact only when the distance between the hand and the object falls within a very close range, leading to poor contact reconstruction performance. Additionally, due to the self-occlusion caused by two-handed interactions and limited viewpoints, HOLD shows poor object reconstruction quality. In contrast, our method effectively addresses the dynamic contact in the ARCTIC dataset by employing $\mathcal{L}_{\text{SDS}}$, which accurately reconstructs occluded parts of the object. Furthermore, we show additional performance improvements for hand reconstruction quality by incorporating the temporal smoothness in $\mathcal{L}_{\text{hand}}$. In Table~\ref{table:table2}, we evaluate the contact reconstruction performance and hand/object reconstruction performance of our method on the HO3D dataset~\cite{hampali2020honnotate}. Additionally, to assess more challenging scenarios, we conduct further evaluations on the HO3D (Limited View) setting, introduced in Section~\ref{sec:exp_datasets}. As a result, while HOLD~\cite{fan2024hold} shows significant performance degradation on the HO3D (Limited View) dataset, our method demonstrates its ability to retain a considerable level of performance from the `Full View' setting, even under the `Limited View' setting. In Table~\ref{table:table3}, we conduct the ablation study on our model by turning on and off 3 losses (\ie $\mathcal{L}_\text{SDS}$, $\mathcal{L}_\text{contact}$ and $\mathcal{L}_\text{smt}$) and hand Gaussian sharing scheme (`share'). When turning off the `share' scheme, we build two separate hand Gaussians. We can see that (1) there is an improvement on hand accuracy (MPJPE) when turning on the `share' scheme and $\mathcal{L}_\text{smt}$, (2) there is an improvement in contact reconstruction accuracy (CD) when turning on $\mathcal{L}_\text{contact}$, (3) there is an improvement in both contact reconstruction (CD) and object reconstruction accuracy (CD$_o$, F10) when turning on $\mathcal{L}_\text{SDS}$. In Table~\ref{table:table4}, we evaluate the rendering quality of our method, using PSNR, SSIM, and LPIPS metrics. This table shows that our 3DGS-based method achieves better rendering quality compared to the NeRF-based alternative~\cite{fan2024hold}.

\noindent \textbf{Qualitative Results.} In Fig.~\ref{fig:fig3-a}, we show the qualitative examples: 3D meshes and 2D rendered images. Since our method estimates only the point clouds for objects, we additionally apply a convex hull algorithm~\cite{convex_hull} to generate object 3D meshes for the visualization. In 3D mesh results, we demonstrate that our method achieves better hand-object alignment in both camera and novel viewpoints. Furthermore, we show our method can effectively reconstruct unseen object surfaces even in the HO3Dv3 (Limited View) setting. In the 2D rendered image results, we demonstrate that our method achieves better rendering quality compared to~\cite{fan2024hold}. Specifically, in the unseen parts of the object, the NeRF-based method~\cite{fan2024hold} shows poor quality; while our results are realistic.
\section{Conclusion}
\label{sec:conclusion}

In this paper, we tackled the challenging task of bimanual category-agnostic interaction reconstruction from monocular videos. We propose a 3D Gaussian splatting (3DGS)-based pipeline for two hands and an unknown object scenario which exhibits severe occlusions. Experiments on ARCTIC dataset demonstrated that our method outperforms recent state-of-the-arts in terms of 3D accuracy and image quality. We further validate the effectiveness of our algorithm in hand-object interaction scenario with limited viewpoints. Ablation studies show that each component works in the meaningful way.

\noindent{\textbf{Limitations and future work:}} Although our method achieves reliable bimanual category-agnostic reconstruction, our reconstruction is limited to articulated hands, where their LBS is known, or rigid objects. Reconstructing articulated unknown objects from a monocular video involves a new challenge and might be an interesting future work.

\noindent{\textbf{Acknowledgements:}} This work is supported by NRF grant (No. RS-2025-00521013 50\%) and IITP grants (No. RS-2020-II201336 Artificial intelligence graduate school program (UNIST) 15\%; No. RS-2021-II212068 AI innovation hub 15\%; No. RS-2022-II220264 Comprehensive video understanding and generation with knowledge-based deep logic neural network 20\%), funded by the Korean government (MSIT).

{
    \small
    \bibliographystyle{ieeenat_fullname}
    \bibliography{main}
}

\end{document}